\expandafter\def\csname ver@fixltx2e.sty\endcsname{}
\documentclass{elsarticle}

\usepackage{url}
\usepackage{microtype}

\usepackage{dblfloatfix}
\usepackage[utf8]{inputenc}
\usepackage{natbib}
\usepackage{amsmath}
\usepackage{amsthm}
\usepackage{amssymb}
\usepackage{amsfonts}
\usepackage{bm}
\usepackage{siunitx}
\usepackage{epstopdf}          
\usepackage{flushend}
\usepackage{parskip}
\usepackage[hyperfootnotes=true,colorlinks=true,linkcolor=blue, bookmarks=true,citecolor={red},urlcolor=cyan]{hyperref}
\usepackage[pagewise]{lineno}
\usepackage{framed} 
\usepackage{multicol} 
\usepackage{tabularx} 
\usepackage{longtable} 
\usepackage[ruled,vlined]{algorithm2e}

\journal{Physics of Fluids}
\usepackage{amssymb}

\usepackage[flushright]{rotating}

\usepackage{subcaption} 
\usepackage[printonlyused]{acronym} 
\usepackage{mathtools} 
\usepackage{comment}

\usepackage[dvipsnames]{xcolor}
\usepackage{tikz}
\usetikzlibrary{fit}
\usetikzlibrary{positioning}
\usetikzlibrary{shapes,arrows}

\definecolor{bluePolimi}{RGB}{22, 44, 80}
\definecolor{lightBluePolimi}{RGB}{91, 122, 172}
\definecolor{greenPolimi}{RGB}{0, 110, 0}
\definecolor{redPolimi}{RGB}{180, 0, 0}

\newcolumntype{M}[1]{>{\centering\arraybackslash}m{#1}}
\usepackage{booktabs} 
\usepackage{comment}
\usepackage{hyperref}

\begin{document}

\newtheorem{theo}{Theorem}
\theoremstyle{definition}
\newtheorem{obs}{Remark}
\newtheorem{Def}{Definition} 
\begin{frontmatter}

\title{Surrogate models for nuclear fusion with parametric Shallow Recurrent Decoder Networks: applications to magnetohydrodynamics}

\author[First]{Matteo Lo Verso}
\author[First]{Carolina Introini}
\author[First]{Eric Cervi}
\author[Second]{Laura Savoldi}
\author[Third]{J. Nathan Kutz}
\author[Fourth,First,cor1] {Antonio Cammi}

\cortext[cor1]{Corresponding author. Email address: antonio.cammi@polimi.it}

\address[First]{Department of Energy, Politecnico di Milano, Milano, 20133, Italy}
\address[Second]{MATHEP Group, Dept. of Energy "Galileo Ferraris", Politecnico di Torino, Torino, Italy}
\address[Third]{Autodesk Research, 6 Agar Street, London UK}
\address[Fourth] {Department of Mechanical and Nuclear Engineering \& Emirates Nuclear Technology Center, Khalifa University, Abu Dhabi, 127788, United Arab Emirates}

\begin{abstract}
    Magnetohydrodynamic (MHD) effects play a key role in the design and operation of nuclear fusion systems, where electrically conducting fluids (such as liquid metals or molten salts in reactor blankets) interact with magnetic fields of varying intensity and orientation, which affect the resulting flow. The numerical resolution of MHD models involves highly nonlinear multiphysics systems of equations and can become computationally expensive, particularly in multi-query, parametric, or real-time contexts. This work investigates a fully data-driven framework for MHD state reconstruction that combines dimensionality reduction via Singular Value Decomposition (SVD) with the SHallow REcurrent Decoder (SHRED), a neural network architecture designed to recover the full spatio-temporal state from sparse time-series measurements of a limited number of observables. The methodology is applied to a parametric MHD test case involving compressible lead-lithium flow in a stepped channel subjected to thermal gradients and magnetic fields spanning a broad range of intensities. To improve efficiency, the full-order dataset is first compressed using SVD, yielding a reduced representation used as reference truth for training. Only temperature measurements from three sensors are provided as input, while the network reconstructs the full fields of velocity, pressure, and temperature. To assess robustness with respect to sensor placement, thirty randomly generated sensor configurations are tested in ensemble mode. Results show that SHRED accurately reconstructs the full MHD state even for magnetic field intensities not included in the training set. These findings demonstrate the potential of SHRED as a computationally efficient surrogate modeling strategy for fusion-relevant multiphysics problems, enabling low-cost state estimation with possible applications in real-time monitoring and control.
\end{abstract}


\end{frontmatter}

\section{Introduction}\label{sec: intro}

Magnetohydrodynamics (MHD) investigates the flow dynamics of electrically conducting fluids under the influence of magnetic fields \cite{freidberg2014ideal}. This theory provides mathematical models extensively used in the nuclear fusion field, especially in magnetic confinement fusion (MCF). Indeed, not only can thermonuclear plasmas be modeled as conducting fluids confined by intense magnetic fields, but MHD theory also applies for the description of the electrically conducting fluids foreseen in the blankets of many tokamaks, like molten salts \cite{ferrero2023impact} or liquid metals \cite{molokov2007liquid}. In fact, in MCF, residual magnetic field lines from the plasma chamber may reach the blanket, interacting with the conducting fluids within and affecting their fluid-dynamics behaviour. Therefore, when designing MCF reactors, MHD effects in the blanket must be considered and properly understood, not only for nominal operations, but also for transient conditions such as plasma disruptions. Given the status of development of MCF systems, numerical investigations of this phenomenon must be adopted.

However, MHD models are systems of nonlinear and highly complex partial differential equations, where the flow and the magnetic field are coupled in a multiphysics framework \cite{biskamp1997nonlinear}. These models require significant computational resources. Additionally, the specific effects induced in the flow by the magnetic field strongly depend on their orientation and intensity \cite{buhler2007liquid}, and simulating every possible case is prohibitive from a computational point of view. The presence of a large number of potential cases becomes even more relevant when it comes to real-time applications for control purposes: in general, high-fidelity models should be able to predict even unforeseen conditions; however, their computational time will likely be too high for any meaningful real-time action. This is a common challenge in multiphysics scenarios governed by nonlinear, strongly coupled sets of equations. In this framework, Reduced Order Modeling (ROM) \cite{lassila2014model, rozza_model_2020} approaches have been studied as a possible strategy to reduce the computational complexity in simulating complex parametric scenarios for engineering applications. Indeed, they provide an efficient alternative to full-order models (FOMs) for multi-query simulations: given a starting high-fidelity dataset, ROM algorithms can construct a surrogate model capable of reproducing the key system physics at a significantly reduced computational cost whilst keeping the desired accuracy. In practice, they project the behavior of the high-dimensional system on a low-dimensional manifold, spanned by the most dominant spatial features, using techniques of dimensionality reduction, including the Singular Value Decomposition (SVD). Once deployed, these surrogate models can operate in quasi-real-time even for previously-unseen parametric configurations and conditions. \newline
As a result, ROM techniques enable rapid exploration of parametric spaces for parametric analysis, uncertainty quantification, and sensitivity, making them particularly suitable for real-time, control-oriented, and design applications in fusion technology. Although data-driven ROM techniques are now well established in many areas of computational physics, including nuclear fission \cite{riva2025data, riva2025real, riva2024impact, riva2024multi, cammi2024data}, their use within MHD physics has only recently begun to emerge \cite{RoyKutz2018_Plasma, kaptanoglu2020characterizing, kaptanoglu2021physics} and remains especially limited for configurations involving electrically conducting liquid metals \cite{loverso2024application, LOVERSO2025115080, loverso_NURETH}.

In parallel with ROM strategies, the fusion community has recently witnessed a rapid growth in the adoption of Machine Learning (ML) and Artificial Intelligence (AI) methodologies, particularly for real-time control, monitoring and digital twin applications \cite{battye2025digital}. Recent approaches rely on deep learning architectures which have been successfully applied to plasma control \cite{degrave2022magnetic}, instability mitigation \cite{seo2024avoiding} and profile regulation \cite{jalalvand2021real} in tokamak devices. These data-driven strategies have demonstrated remarkable capabilities in learning highly nonlinear dynamics and in enabling fast predictions. However, despite their promising performance, purely data-driven AI typically require very large training datasets and entail substantial training times, which may become prohibitive when high-fidelity simulations are expensive or when experimental data are scarce, noisy or difficult to acquire. These limitations are particularly critical in MHD scenarios involving liquid metal flows in fusion blankets, where generating extensive datasets under realistic operating conditions remains a major challenge.

In this context, an appealing alternative consists in exploiting ML techniques within a reduced and physically informed framework, where the dimensionality of the problem is first compressed by reduced order modelling techniques. By performing learning in a low-dimensional latent space, it is possible to significantly reduce the amount of training data and computational effort required, while retaining the essential physical features of the underlying MHD dynamics. This compressive training paradigm provides a natural bridge between physics-based modelling and data-driven approaches, and represents a particularly suitable strategy for MHD applications in fusion technology. By compressing the starting dataset, the training cost of ML models can be significantly reduced, compared to training directly in the high-dimensional space. Moreover, this framework facilitates the integration of measurements collected from the physical system with prior knowledge from models, offering advantages over conventional data assimilation techniques, which, being based on optimization problems, are often limited by long computational times. \newline
Within this framework, this work discusses the possibility of adopting a combination of SVD and an ML technique to provide an accurate and reliable state reconstruction of MHD physics, considering a parametric scenario. The selected architecture is the SHallow REcurrent Decoder (SHRED) \cite{williams2024sensing, faraji2025shallow}, an ML architecture capable of mapping sparse trajectories of a measured observable to the full high-dimensional state space, thereby indirectly estimating also unmeasured quantities. Through a recurrent unit followed by a shallow decoder, SHRED efficiently learns the spatio-temporal dynamics of the system, even when trained with a small number of sensors. More importantly, it can generalize across different parameter values, making it suitable in the MHD framework for reconstructing flows under a range of various magnetic field intensities and orientations. This work represents the first application of the SHRED methodology to MHD physics for conducting fluids: to assess its performance for this class of problems, the selected test case is a compressible MHD flow in a channel with steps and thermal gradients.

The structure of the present paper is now reported. Section \ref{sec: SHRED} provides an overview of the SHRED architecture. Section \ref{sec: Numerical results} describes the MHD model and presents the key numerical results. Finally, Section \ref{sec: Conclusions} resumes the main conclusions of the present work, along with some future perspectives.

\section{SHallow REcurrent Decoder}\label{sec: SHRED}

\begin{figure}[htbp]
    \centering
    \includegraphics[width=1.0\linewidth]{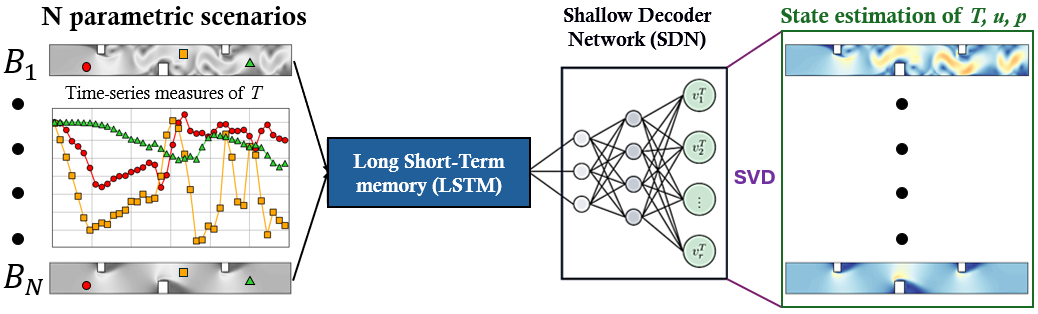}
    \caption{Visual sketch of the SHRED architecture applied to the MHD channel flow. Following compression of the starting dataset through SVD, three sensors are used for measuring the local evolution of temperature over time. The temporal trajectories are encoded in a latent space through a long-short-term memory (LSTM). Then, a Shallow Decoder Network (SDN) projects the resulting latent representation into a compressive representation of all spatio-temporal field variables. Finally, the compressive representation is mapped back to the full-order state space by the SVD.}
    \label{fig: SHRED_architecture}
\end{figure}

The SHallow REcurrent Decoder network (SHRED) is a novel and promising data-driven machine learning technique first proposed by \cite{williams2024sensing}, designed for state estimation and forecasting of complex dynamical systems from sparse time-series measurements. Its standard architecture consists of a Long Short-Term Memory (LSTM) unit \cite{LSTM2016long} to capture temporal dependencies in the latent dynamics and a Shallow Decoder Network (SDN) \cite{SDN20shallow} for nonlinear mapping between latent and physical spaces. In this work, a compressed version of SHRED is exploited: the training dataset is pre-processed by compression through Singular Value Decomposition (SVD), significantly reducing the number of features\footnote{It must be mentioned that, in this case, the reference truth becomes the SVD compression, which acts as a lower bound for the reconstruction error of the starting dataset.}. The use of SVD within SHRED has been shown to significantly enhance computational efficiency by reducing the dimensionality of the data at the training level \cite{tomasetto2025reduced}, allowing for training even on personal laptops. Figure \ref{fig: SHRED_architecture} shows the architecture of the SHRED network used in this work.

At first, the architecture learns the temporal evolution of the system trajectories in accordance with Takens' embedding theory \cite{takens2006detecting}, which states that the dynamics of a high-dimensional system can be reconstructed from a sequence of time-delayed observations of a few variables: the LSTM captures the temporal dependencies and nonlinear correlations embedded within the sensor measurements. Subsequently, the SDN maps the learned latent trajectories back to the reconstructed space, where SVD is employed to decompress the latent features and recover the full-state representation of the system.

This architecture offers several advantages over traditional data-driven techniques for surrogate models. First, SHRED has been proven to be able to perform accurate state reconstructions with an exceptionally small number of sensors (typically three are enough) beyond which reconstruction errors tend to saturate \cite{tomasetto2025reduced}: this property makes SHRED particularly effective in low-measurement or high-cost sensing scenarios. Furthermore, SHRED is agnostic to sensor locations, since it has been proven to achieve accurate state reconstruction even when sensors are randomly distributed \cite{williams2024sensing}: this means that sensors can be placed where installation is most practical or accessible, and that optimization of the sensor positioning is no longer a hard requirement. In fusion systems, where the placement of sensors may be constrained by geometry, temperature, and radiation conditions, this is an important benefit, since  SHRED can provide a practical and efficient strategy to reconstruct the entire dynamical field from a minimal set of available measurements, located in easy-to-access parts of the domain. Furthermore, the model can process multiphysics data derived from a single observable, enabling the recovery of strongly coupled quantities of interest even when direct measurements are unavailable. This capability may be especially beneficial in tokamak systems, where certain quantities (such as temperature) are easier to measure with respect to others (like fluid velocity, neutron flux). By exploiting correlations learned during training, SHRED can represent a strategy for estimating all the variables of interest from the most accessible signals. Compared to other ML techniques, SHRED can be trained directly on compressed data representations, greatly reducing computational costs and memory usage, allowing laptop-level training without the need for high-performance computing. Additionally, SHRED requires minimal hyperparameter tuning, as it has been proven that the same architecture works efficiently across very diverse physical systems \cite{williams2024sensing}. 
A further key advantage of SHRED, which is particularly relevant in nuclear engineering, lies in its strong mathematical foundations. The methodology builds upon Takens’ embedding theorem \cite{takens2006detecting} and can be interpreted as a generalization of classical separation of variables approach to data-driven settings. This theoretical foundation, combined with the shallow network architecture, results in a model with a very limited number of trainable parameters (typically fenwer than $10^3$), which stands in contrast to many deep learning approaches relying on millions of parameters. As a consequence, SHRED offers a higher degree of interpretability, facilitating physical insight into the learned dynamics and increasing confidence in its application to safety-relevant nuclear scenarios. All these features make SHRED an excellent candidate for state reconstruction in complex physics. So far, SHRED has been successfully tested across a wide range of physical systems \cite{williams2024sensing, faraji2025shallow, tomasetto2025reduced, moen2025mapping}, consistently demonstrating excellent performance and generalization capabilities. In the context of nuclear applications, SHRED has been previously employed for state estimation in scenarios involving fission reactors \cite{riva2025constrained, riva2025robust, riva2025towards, introini2025models}, but it has never been applied to fusion systems. 

In its original formulation, the SHRED was proposed in a single-parameter configuration \cite{williams2024sensing, faraji2025shallow,riva2025robust}, focusing on the reconstruction of system dynamics under fixed physical conditions. However, the same architecture can be easily extended to parametric datasets, as done in \cite{tomasetto2025reduced,riva2025towards}. This flexibility arises from the intrinsic design of SHRED: since the LSTM operates on lagged time-series data, the architecture naturally accommodates multiple trajectories corresponding to different parameter values. In this extended setting, a physical parameter $\mu$ can be incorporated either as an additional input, when its value is known, or as an output variable, when parameter estimation is desired. 

More in detail, for each parameter value $\boldsymbol{\mu}_p$, the snapshot matrix is defined as $\mathbb{X}^{\boldsymbol{\mu}_p} \in \mathbb{R}^{\mathcal{N}_h \times N_t}$, where $\mathcal{N}_h$ denotes the number of cells of the mesh (number of features) and $N_t$ the number of saved time instants. Then, the resulting matrix is compressed with an SVD through the reduced basis $\mathbb{U}^{\boldsymbol{\mu}_p} \in \mathbb{R}^{\mathcal{N}_h \times r}$ of rank $r$, from which a corresponding latent representation $$
\mathbb{V}^{\boldsymbol{\mu}_p}=\left(\mathbb{U}^{\boldsymbol{\mu}_p}\right)^T \mathbb{X}^{\boldsymbol{\mu}_p} \in \mathbb{R}^{r \times N_t}$$ can be derived. $\mathbb{V}^{\boldsymbol{\mu_p}}$ represents the temporal coefficients which embed the dynamics associated with the parameter $\mu_p$, and are used as training data for SHRED. However, when dealing with a parametric dataset, it is necessary to construct a common reduced basis that spans the entire parametric space, thus encoding the most representative physical features across different parameter values. Then, the full dataset is stacked in the form:
\begin{equation}
    \mathbb{X}=\left[\mathbb{X}^{\boldsymbol{\mu}_1}\left|\mathbb{X}^{\boldsymbol{\mu}_2}\right| \ldots \mid \mathbb{X}^{\boldsymbol{\mu}_{N_p}}\right]
\end{equation}
where $N_p$ represents the number of instances of the parameter.

In the present work, the parameter of interest is represented by the intensity of the applied vertical magnetic field, which plays a crucial role in determining the evolution of the lead–lithium flow within the fusion reactor blanket \cite{buhler2007liquid, muller2001magnetofluiddynamics}. Extending SHRED to this parametric configuration enables the model to learn how changes in the magnetic field affect the flow dynamics, thereby providing a powerful data-driven tool for studying MHD phenomena in fusion environments. More generally, the proposed framework is not limited to this specific choice and can be easily extended to other relevant parameters (such as the inlet velocity or the orientation of the applied magnetic field). The focus on a vertically applied magnetic field in this study is motivated by practical considerations. For the sake of simplicity, restricting the analysis to a single parametric direction allows for a clearer assessment of the capability of SHRED to generalise across different operating conditions, while limiting the complexity of the training dataset. Extending the approach to multi-parameter spaces, including arbitrary magnetic field orientations and flow conditions, is therefore a natural and feasible direction for future investigations.

The SHRED architecture has been implemented in Python utilizing the PyTorch package, adapting the original code of \cite{williams2024sensing}. Both the LSTM and the SDN units of the implemented SHRED architecture are composed of $2$ hidden layers: the layers of the former have $64$ neurons each, whereas those of the latter consist of $350$ and $400$ neurons, respectively.

\section{Numerical Results} \label{sec: Numerical results}

\begin{figure}[htbp]
    \centering
    \includegraphics[width=1.0\linewidth]{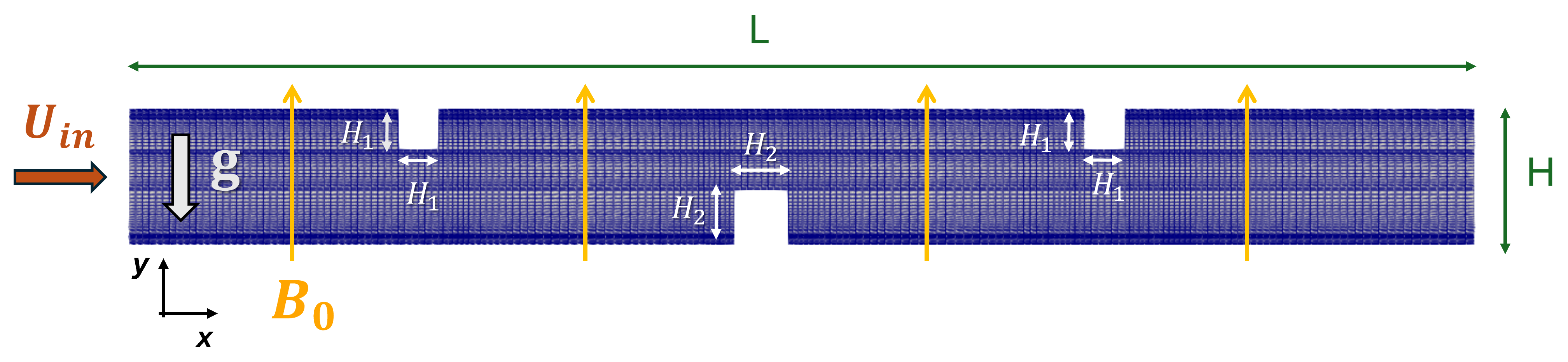}
    \caption{Computational domain of the selected benchmark}
    \label{fig: geometry}
\end{figure}

The selected test case, shown in Figure \ref{fig: geometry}, consists of lead-lithium MHD flow in a bi-dimensional channel with multiple steps. Although the selected benchmark does not correspond to any specific blanket geometry, it provides an interesting test case for a first application of SHRED to MHD physics for several reasons: first and foremost, despite its apparent simplicity, this setup retains all the key MHD phenomena relevant to liquid metal flows in fusion blankets, while involving a sufficiently intricate multiphysics coupling.

The geometry includes two steps on the upper wall and one on the lower wall, representing obstacles to the flow. The upper steps are assumed at a temperature lower than the inlet fluid temperature $T_0$, while the bottom step is set at a higher temperature. These temperature conditions produce thermal gradients in the flow and, consequently, density variations and potential buoyancy effects superimposed on the main flow.  In addition, the three steps act as physical obstacles that, in the absence of a magnetic field, would produce strong turbulent dynamics. However, when a magnetic field is imposed, the resulting Lorentz force suppresses the small-scale motions, leading to a progressive laminarization and regularization of the flow \cite{lee2001magnetohydrodynamic}: the level of laminarization depends directly on the magnetic field intensity. Although this scenario does not directly represent a realistic blanket geometry, it constitutes a meaningful test case for evaluating the ability of SHRED to accurately reconstruct complex flow dynamics. In particular, it allows the assessment of how the technique captures the varying degrees of turbulence suppression and convective effects that arise in MHD flows depending on the intensity of the magnetic field.

As an initial condition, the flow is assumed to be at null velocity, and a perpendicular magnetic field $\mathbf{B}_0$ is imposed in the domain. Regarding boundary conditions, a uniform fluid velocity at the inlet and an external pressure at the outlet are imposed. Moreover, all the walls are assumed to be no-slip and perfectly electrically conducting, subjected to the uniform vertical magnetic field $\mathbf{B}_0$. The resulting magnetohydrodynamic model for the considered compressible, visco-resistive MHD flow \cite{pu2013global, tang2015ideal} is the following:
\begin{equation}
    \left\{\begin{array}{ll}
    \dfrac{\partial \rho}{\partial t} +  \nabla \cdot \left( \rho \mathbf{u}\right) = 0 & \text{in } \Omega, \, t > 0  \\[5pt]
    \dfrac{\partial (\rho \mathbf{u}) }{\partial t} + \nabla \cdot \left( \rho \mathbf{u} \otimes \mathbf{u} \right) = - \nabla p + \nabla \cdot \boldsymbol{\tau} + \rho \mathbf{g} + \left( \dfrac{1}{\mu_0} \nabla \times  \mathbf{B}\right) \times \mathbf{B} & \text{in } \Omega, \, t > 0 \\[1pt]
    \dfrac{\partial (\rho c_v T) }{\partial t} + \nabla \cdot (\rho c_v T  \mathbf{u})  =  \kappa \Delta T + \dfrac{1}{\sigma \mu_0^2}|\nabla \times \boldsymbol{B}|^2 & \text{in } \Omega, \, t > 0 \\[1pt]
    \rho = \rho_0 \left(1 - \beta\left(T - T_0 \right) \right) & \text{in } \Omega, \, t > 0\\[1pt]
    \boldsymbol{\tau} =\mu\left(\nabla \mathbf{u}+(\nabla \mathbf{u})^T\right)-\dfrac{2}{3} \mu(\nabla \cdot \mathbf{u} \, \mathbf{I}) & \text{in } \Omega, \, t > 0\\[6pt]
    \dfrac{\partial{ \mathbf{B}}}{\partial{t}} = \nabla \times (\mathbf{u} \times \mathbf{B}) + \dfrac{1}{ \sigma \mu_0}  \Delta \mathbf{B} & \text{in } \Omega, \, t > 0 \\[6pt]
    \nabla \cdot \mathbf{B} = 0 & \text{in } \Omega, \, t > 0 \\
    \end{array}\right.
\end{equation}
with the following initial and boundary conditions: 
\begin{equation}
    \left\{\begin{array}{ll}
    \mathbf{u} = \mathbf{0} , \, T = T_0, \, \rho = \rho_0, \, \mathbf{B} = \mathbf{B}_0 & \text{in } \Omega, \, t = 0 \\ 
    \mathbf{u} = \mathbf{u}_{in} & \text{on } \Gamma_{inlet},  \, t > 0 \\
    \frac{\partial \mathbf{u}}{\partial \mathbf{n}}  = 0 & \text{on } \Gamma_{outlet},  \, t > 0  \\
    \mathbf{u} = \mathbf{0} & \text{on } \Gamma_{walls},  \, t > 0 \\
    \mathbf{B} = \mathbf{B}_0 & \text{on } \Gamma_{walls},  \, t > 0   \\
    \frac{\partial \mathbf{B}}{\partial \mathbf{n}} = 0 & \text{on } \Gamma_{inlet} \cup \Gamma_{outlet},  \, t > 0   \\
    p = p_{out} & \text{on } \Gamma_{outlet}, \, t > 0 \\
    \frac{\partial p}{\partial \mathbf{n}} = 0 & \text{on } \partial \Omega \backslash \Gamma_{outlet}, \, t > 0 \\
    T = T_{top} & \text{on } \Gamma_{top \, steps}, \, t > 0 \\
    T = T_{bottom} & \text{on } \Gamma_{bottom \, step}, \, t > 0 \\
    \end{array}\right.
\end{equation}
where $\Omega$ represents the domain, $\partial \Omega$ the entire boundary, $\Gamma$ the surfaces of the boundary and $t$ is the time. Moreover, $\mathbf{u}$ is the fluid velocity,  $p$ the pressure, $\mathbf{B}$ the magnetic field, $\rho$ the density, $\boldsymbol{\tau}$ the viscous stress tensor, $T$ the temperature, $\mathbf{g}$ the gravity, $\mu$ the dynamic viscosity,  $\mu_0$ the magnetic permeability and $\sigma$ the electrical conductivity, $\kappa$ the thermal conductivity, $c_v$ the specific heat. All physical and numerical parameters, including the initial and boundary conditions, are reported in Table \ref{tab: parameters}. The proposed model consists of a complex system of equations featuring strong multiphysics coupling: the fluid variables (velocity, pressure, and temperature) are mutually dependent and are also influenced by the specific magnetic field experienced by the fluid. 

\renewcommand{\arraystretch}{1.1}
\begin{table}[htbp]
    \centering
    \caption{Physical and numerical parameters for the FOM.}
    \label{tab: parameters}
    \begin{tabular}{|c|c|c|c|c|c|c|c|c|} \hline
    $\rho_0$ & $9806\ \text{kg/m}^3$ & & $\kappa$ & $20.93\ \text{Wm}^{-1}\ \text{K}^{-1}$ & & $\mathcal{N}_h$ & $14460$ \\ \hline
    $\mu$ & $1.93*10^{-3}\ \text{Pa} \cdot \text{s}$ & &  $u_{in}$ & $0.0492\ \text{m/s}$ & & $L$ & $0.2\ \text{m}$ \\ \hline
    $\mu_B$ & $1.26 *10^{-6} \ \text{H/m}$ & & $p_{out}$ & $10^5\ \text{Pa}$ & & $H$ & $0.02\ \text{m}$ \\ \hline
    $\sigma$ & $7.82*10^5\ \Omega^{-1} \text{m}^{-1}$ & & $T_0$ & $600\ \text{K}$ & & $H_1$ & $0.006\ \text{m}$\\ \hline
    $\beta$ & $1.3*10^{-4}\ \text{K}^{-1}$ & & $T_{top}$ & $550\ \text{K}$ & & $H_2$ & $0.008\ \text{m}$\\ \hline
    $c$ & $189.5\ \text{J}\text{Kg}^{-1}\text{K}^{-1}$ & & $T_{bottom}$ & $650\ \text{K}$ & & $z$ & $10^{-4}\ \text{m}$ (1 cell)\\ \hline
\end{tabular}\end{table}
\renewcommand{\arraystretch}{1}

In this analysis, $N_p = 19$ different values for the magnetic field intensity are considered. The selected MHD scenario has been solved numerically multiple times, imposing different values for the magnetic field in a range between $0.01 \;\text{T}$  and $0.5 \;\text{T}$. Each considered case has been simulated up to $3 \,s$ with a variable timestep according to the $CFL$ condition, to ensure numerical stability. Data were saved every $0.025 \, s$ (so $N_t = 120)$. All the snapshots have been generated using the OpenFOAM MHD library \textbf{magnetoHDFoam}, developed in \cite{loverso2024solver} and available on \url{https://github.com/ERMETE-Lab/MHD-magnetoHDFoam} under the MIT license. The snapshots simulations have been performed on an HPC cluster, with each case requiring approximately $20$ minutes. 

\begin{figure}[htbp]
    \centering
    \includegraphics[width=1\linewidth]{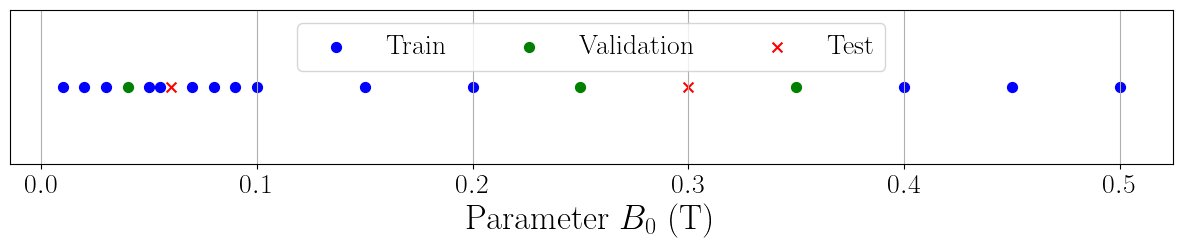}
    \caption{Subdivision of the dataset in training, validation and test snapshots.}
    \label{fig: range_splitting}
\end{figure}

\begin{figure}[htbp]
    \centering
    \includegraphics[width=1\linewidth]{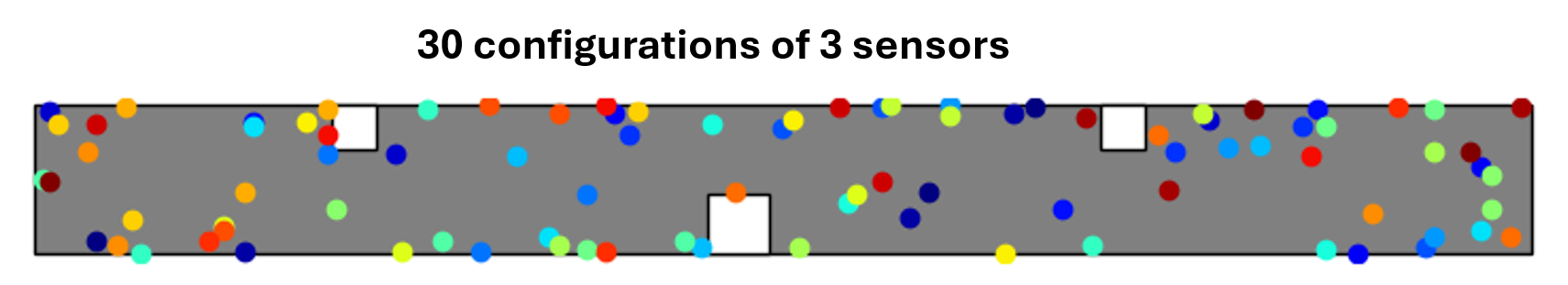}
    \caption{Visualization of the $30$ randomly generated configurations of triplets of sensors used in this work for recording point measurements of temperature dynamics. Each color corresponds to a different triplets of sensors.}
    \label{fig: sensors_configurations}
\end{figure}

The snapshots of each field have been stacked as described in Section \ref{sec: SHRED}, and they have been rescaled using the min-max formula\footnote{A common practice in Machine Learning to improve the efficiency and performance of the model.}, i.e.:
\begin{equation}
    \tilde{T}= \frac{T - T_{min}}{T_{max} - T_{min}}, \quad
    \tilde{\mathbf{u}}= \frac{\mathbf{u} - \mathbf{u}_{min}}{\mathbf{u}_{max} - \mathbf{u}_{min}}, \quad\tilde{p}= \frac{p' - p'_{min}}{p'_{max} - p'_{min}}
\end{equation}
where $p' = p-\rho gh$ represents the pressure without the hydrostatic component. In the following, all variables will be considered in their normalized form, and for simplicity of notation, they will be denoted simply as $T$, $\mathbf{u}$, and $p$. The scaled dataset has been divided into training ($\simeq73.7\%$), validation ($\simeq15.8\%$), and test ($\simeq10.5\%$) snapshots. This subdivision follows a standard practice in machine learning, where approximately three quarters of the available data are used for training, while the remaining portion is reserved for validation and testing. In this framework, the surrogate model is trained using only a subset of the dataset, and its accuracy is subsequently assessed by comparing the surrogate model predictions with the test data, which are not seen during the training phase. \newline
Figure \ref{fig: range_splitting} reports an overview of the performed subdivision of the data. 
It can be observed that the dataset is denser for lower magnetic field intensities and sparser for higher ones. This choice is motivated by the fact that, as previously discussed, the magnetic field tends to laminarize the flow, and higher magnetic field intensities generally lead to more homogeneous and stable flow patterns \cite{lee2001magnetohydrodynamic}. Consequently, it is more appropriate to enrich the dataset with cases characterized by lower magnetic field strengths, where the dynamics are more complex, variable, and diverse, and therefore more informative for training the model. Two different test cases have been selected, one associated with a very low ($B_0= 0.06 \, T$) and one with a quite high ($B_0= 0.3 \, T$) magnetic field intensity. This selection has been done to test the ability of the SHRED to reconstruct MHD scenarios subjected to both weak and strong vertical magnetic fields, and thus to retrieve a general representation even considering different dynamics.
As previously explained, SHRED requires a limited set of time-series measures of a single field to establish a mapping between the observed values of that field and the reduced coefficients of all fields. To this end, several sensors were placed within the geometry to collect the measurements of the temperature field. Notably, SHRED is able to operate effectively with only three sensors, as shown in \cite{tomasetto2025reduced}. However, to verify its independence from the locations of sensors, $30$ randomly distributed triplets of sensors were considered (see Figure \ref{fig: sensors_configurations}), building $30$ distinct SHRED models, each associated with a different configuration (ensemble mode). To numerically generate the sensor measurements, the temperature values over time corresponding to the mesh cells associated with each sensor location were extracted from the dataset.

\begin{figure}[htbp]
    \centering
    \includegraphics[width=1\linewidth]{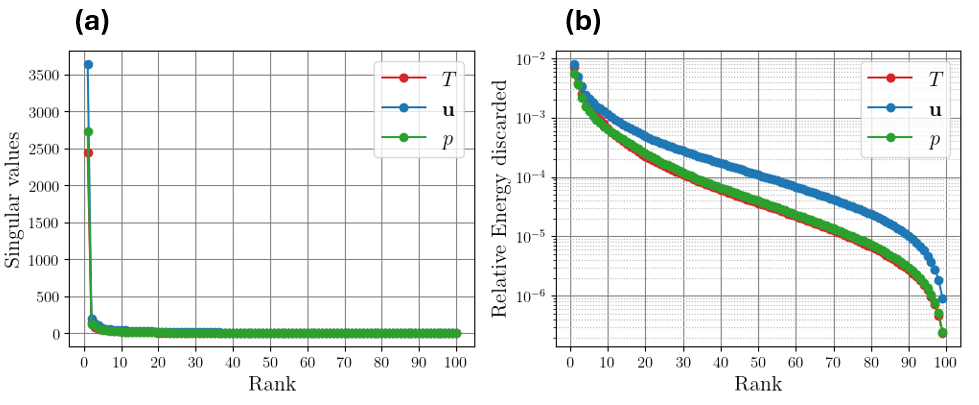}
    \caption{Singular values (a) and relative information/energy content discarded (b) of the training snapshots as a function of the rank for the temperature, velocity and pressure fields.}
    \label{fig: singular_values}
\end{figure}

The dimensionality of the snapshots is now reduced through the SVD, building a reduced representation of the problem considering only the first $r$ principal modes. To select the rank $r$ of the reduced space, the decay of the singular values related to the training snapshots has been investigated. Figure \ref{fig: singular_values} shows both the decay of singular values and the relative information/energy discarded as a function of $r$. By examining the decay of the singular values and the associated relative information, a rank of $r=20$ was selected. This choice ensures that only a negligible portion of the total information is neglected (less than $0.1 \%$), ensuring that the reduced representation still encodes not only the dominant large-scale behavior but also the relevant small-scale dynamics. 

\begin{figure}[h]
    \centering
    \includegraphics[width=1\linewidth]{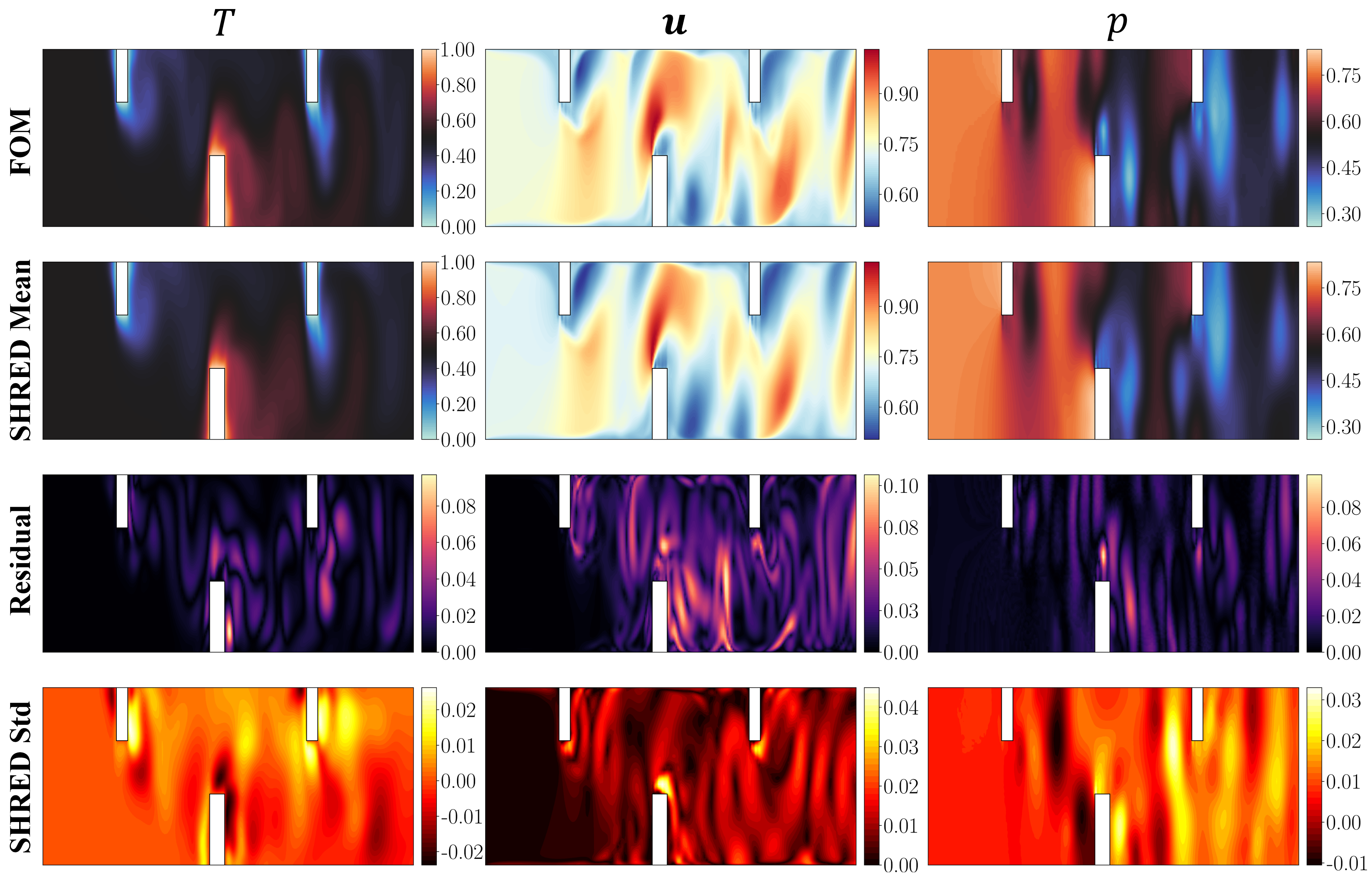}
    \caption{Results for the temperature (first column), velocity (second column) and pressure (third column) for the case with $B_0=0.06 \, T$ at time $t=2\, s$. The first row displays the reference full-order solution while the second row shows the mean reconstruction obtained by averaging the outputs of the $30$ SHRED models. The third row reports the difference between the FOM and the mean SHRED reconstruction while the fourth row shows the standard deviation among the $30$ reconstructions.}
    \label{fig: reconstruction_Blow}
\end{figure}

\begin{figure}[h]
    \centering
    \includegraphics[width=1\linewidth]{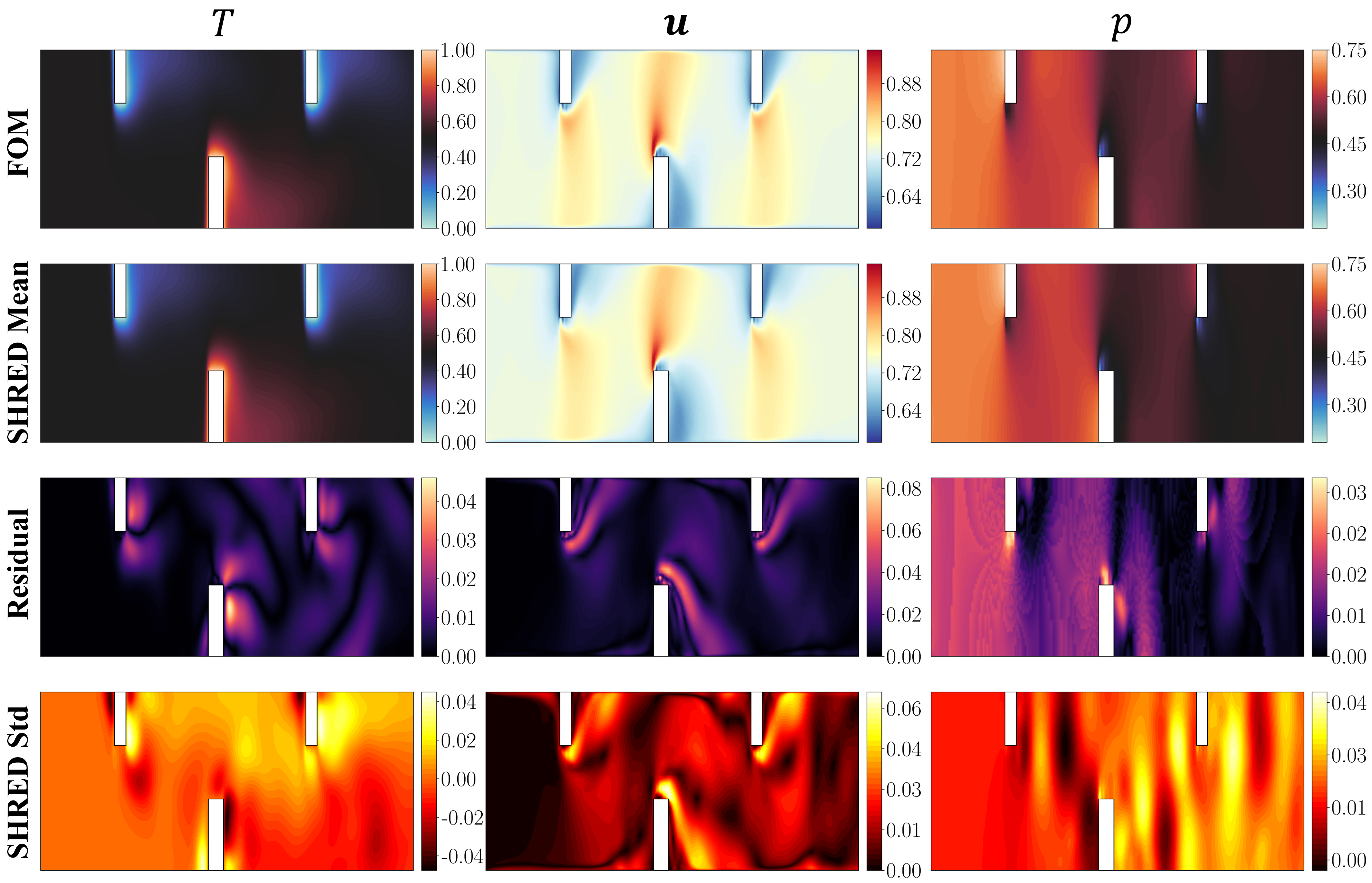}
    \caption{Results for the temperature (first column), velocity (second column) and pressure (third column) for the case with $B_0=0.3 \, T$ at time $t=2\, s$. The first row displays the reference full-order solution while the second row shows the mean reconstruction obtained by averaging the outputs of the $30$ SHRED models. The third row reports the difference between the FOM and the mean SHRED reconstruction while the fourth row shows the standard deviation among the $30$ reconstructions.}
    \label{fig: reconstruction_Bhigh}
\end{figure}

During the training phase, SHRED was trained using the temperature measurements and the compressed representations of the training and validation snapshots in order to learn a mapping between the sensor inputs and the corresponding SVD temporal coefficients. Each SHRED model took about $10$ minutes for the training phase on a personal computer with an Intel Core i7-9800X processor. Subsequently, in the test phase, SHRED takes as input only the temperature measurements from the test case and, using the mapping between measures and SVD coefficients learned during training, reconstructs the full state for the (unseen) value of the magnetic field intensity. The associated computational time required from each trained SHRED to generate the new output is practically null (less than $1 \, s$). At this point, the outputs of the $30$ models were averaged, and the mean result is taken into account. \newline
Figure \ref{fig: reconstruction_Blow} shows the results obtained for the test case with the lower magnetic field intensity. In particular, the truth solution, corresponding to the effective numerical resolution of the full-order model, and the average SHRED reconstruction are compared. The comparison shows that the SHRED model is able to reproduce the evolution of all the considered fields with remarkable accuracy, relying solely on temperature field measurements. The reconstructed solution closely matches the full-order one, and the residuals (computed as the absolute difference between the FOM and the SHRED) are generally very small, with noticeable values only at a few regions located after the steps, where the dynamics are more complex. \newline
Figure \ref{fig: reconstruction_Bhigh} reports the results obtained in the test case with the higher magnetic field. The results show that, under a stronger magnetic field, SHRED exhibits an even enhanced ability to capture the dynamics of the relevant fields. As illustrated in the figure, the reconstructed solution closely matches the original one, and the residuals are even lower than in the previous case. This improvement arises because higher magnetic field intensities tend to further suppress vortical dynamics through the Lorentz force, promoting a completely laminarised and more homogeneous flow, which is easier to reconstruct, as the small-scale chaotic structures are damped. \newline
Moreover, a comparison between Figures \ref{fig: reconstruction_Blow} and \ref{fig: reconstruction_Bhigh} clearly puts in evidence that the MHD dynamics strongly depend on the specific value of the magnetic field, as the flows obtained in the two considered cases are completely different. However, a single SHRED model, trained over a broad range of magnetic field intensities, is capable of accurately reconstructing both physical scenarios. This demonstrates that the architecture can generalize across highly distinct physical regimes, capturing the underlying dynamics even when the input conditions vary significantly. \newline
Furthermore, Figures \ref{fig: reconstruction_Blow} and \ref{fig: reconstruction_Bhigh} show the standard deviations across the outputs of the $30$ SHRED models. The computed deviations are consistently low throughout the entire geometry for both the considered test cases. This indicates that the $30$ solutions, each corresponding to a different configuration of three sensor locations, are highly similar, with differences that are practically negligible, further proving the agnosticism of SHRED to sensor locations.

The results presented so far illustrate the reconstructed flow fields across the entire geometry at a fixed time instant. To further demonstrate the accuracy of SHRED models over the whole temporal window, the time evolution of selected global quantities is analyzed. For each physical field, the temporal evolution of its spatial average across the geometry is analyzed. As already explained, for every time instant, the $30$ trained SHRED models provide $30$ reconstructions over the domain. These outputs were first averaged across the $30$ models to obtain a representative mean reconstruction at each time step, which was shown previously. Now, a further spatial average over the entire geometry is computed from this mean reconstruction, yielding a single time-dependent quantity that characterizes the overall evolution of the field.


\begin{figure}[h]
    \centering
    \includegraphics[width=1\linewidth]{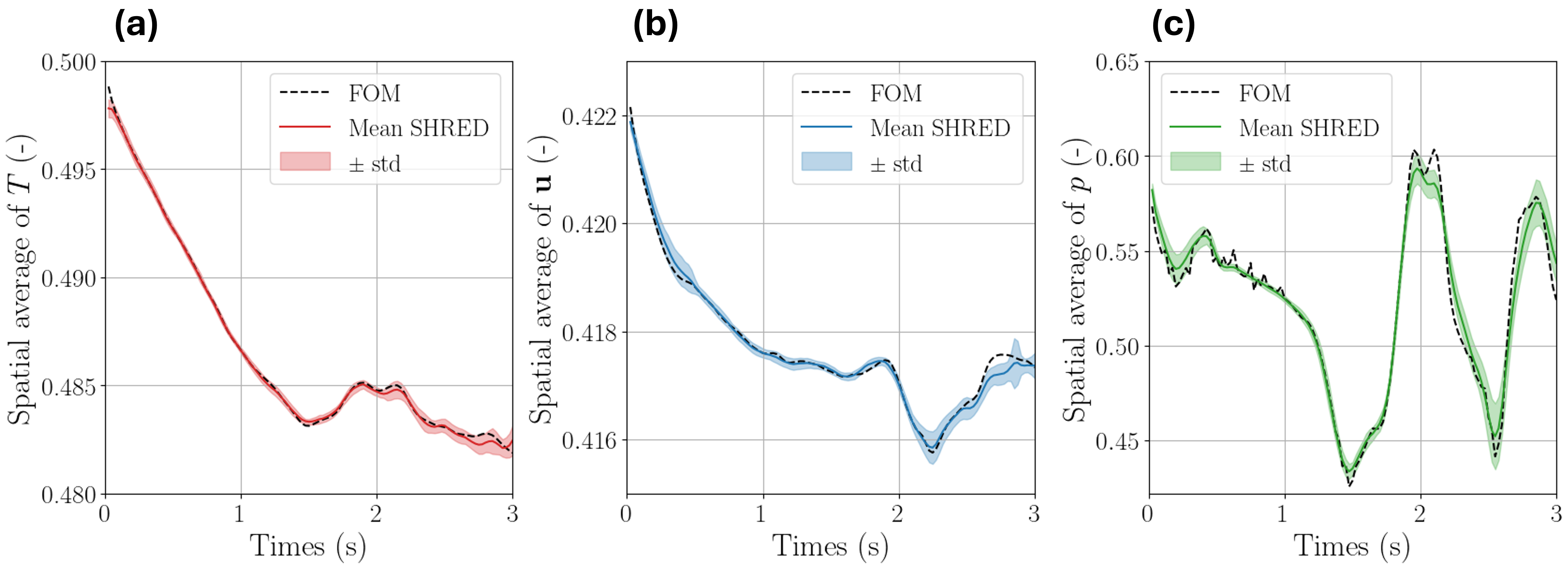}
    \caption{Temporal evolution of the spatial averages across the geometry for the temperature (a), velocity (b) and pressure (c) fields in the test case with $B_0 = 0.06 \, T$. For each field the true profile associated to the full-order solution is compared with the average across the $30$ reconstructions obtained, with the related standard deviations.}
    \label{fig: mean_field_lowB}
\end{figure}

\begin{figure}[h]
    \centering
    \includegraphics[width=1\linewidth]{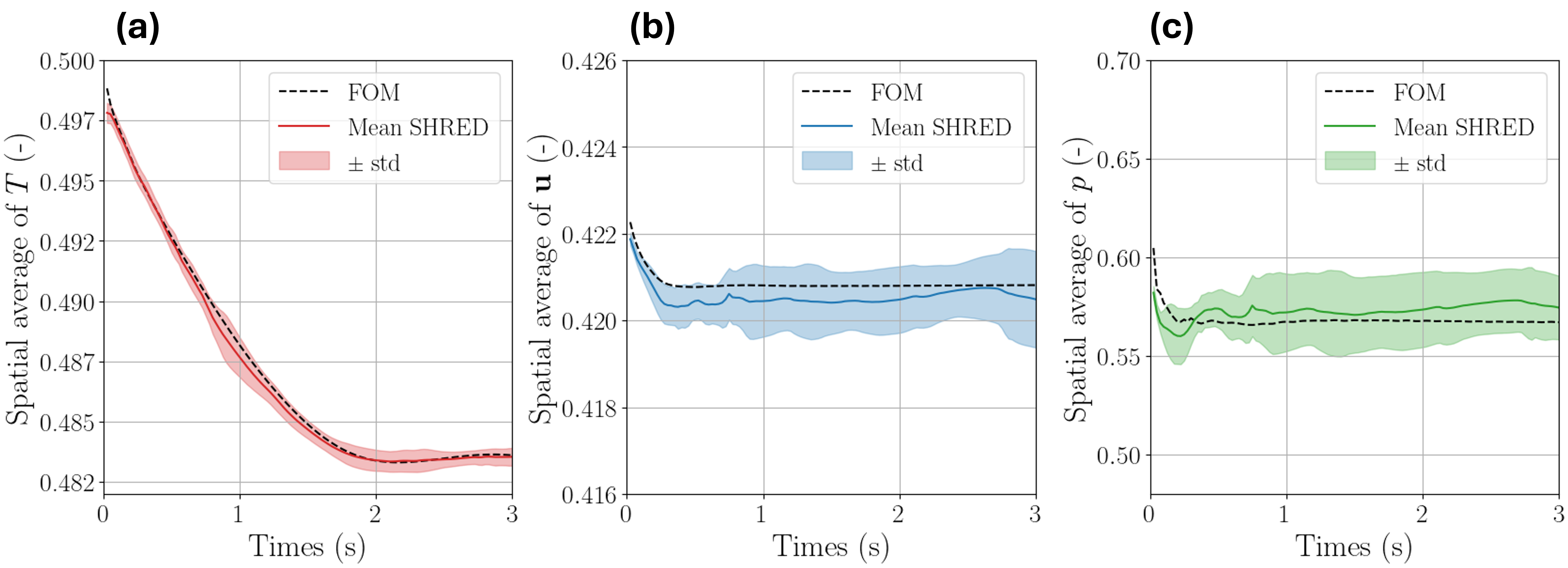}
    \caption{Temporal evolution of the spatial averages across the geometry for the temperature (a), velocity (b) and pressure (c) fields in the test case with $B_0 = 0.3 \, T$. For each field the true profile associated to the full-order solution is compared with the average across the $30$ reconstructions obtained, with the related standard deviations.}
    \label{fig: mean_field_highB}
\end{figure}

In addition, to assess the consistency among the models, the standard deviation of the spatial averages across the $30$ reconstructions is also calculated. This allows evaluating not only the accuracy of the mean reconstruction with respect to the full-order model dynamics, but also how similar the outputs of the different sensor configurations are in terms of their global temporal behavior. Figures \ref{fig: mean_field_lowB} and \ref{fig: mean_field_highB} report the temporal dynamics of the spatially averaged temperature, velocity, and pressure of the fluid for the cases with lower and higher magnetic field, respectively.

The results clearly show that the SHRED reconstruction closely follows the full-order profile for all the fields and throughout the entire time interval, confirming the SHRED capability to accurately reconstruct the true flow dynamics. Moreover, standard deviations remain low for all reconstructed fields over time, indicating that the outputs of the $30$ models are highly similar, despite being trained on input temperature measurements taken at different sensor locations. This further confirms that SHRED is effectively agnostic to sensor placement, meaning that its reconstruction performance does not depend on the specific positions of the sensors providing the input data. 

\begin{figure}[htbp]
    \centering
    \includegraphics[width=1\linewidth]{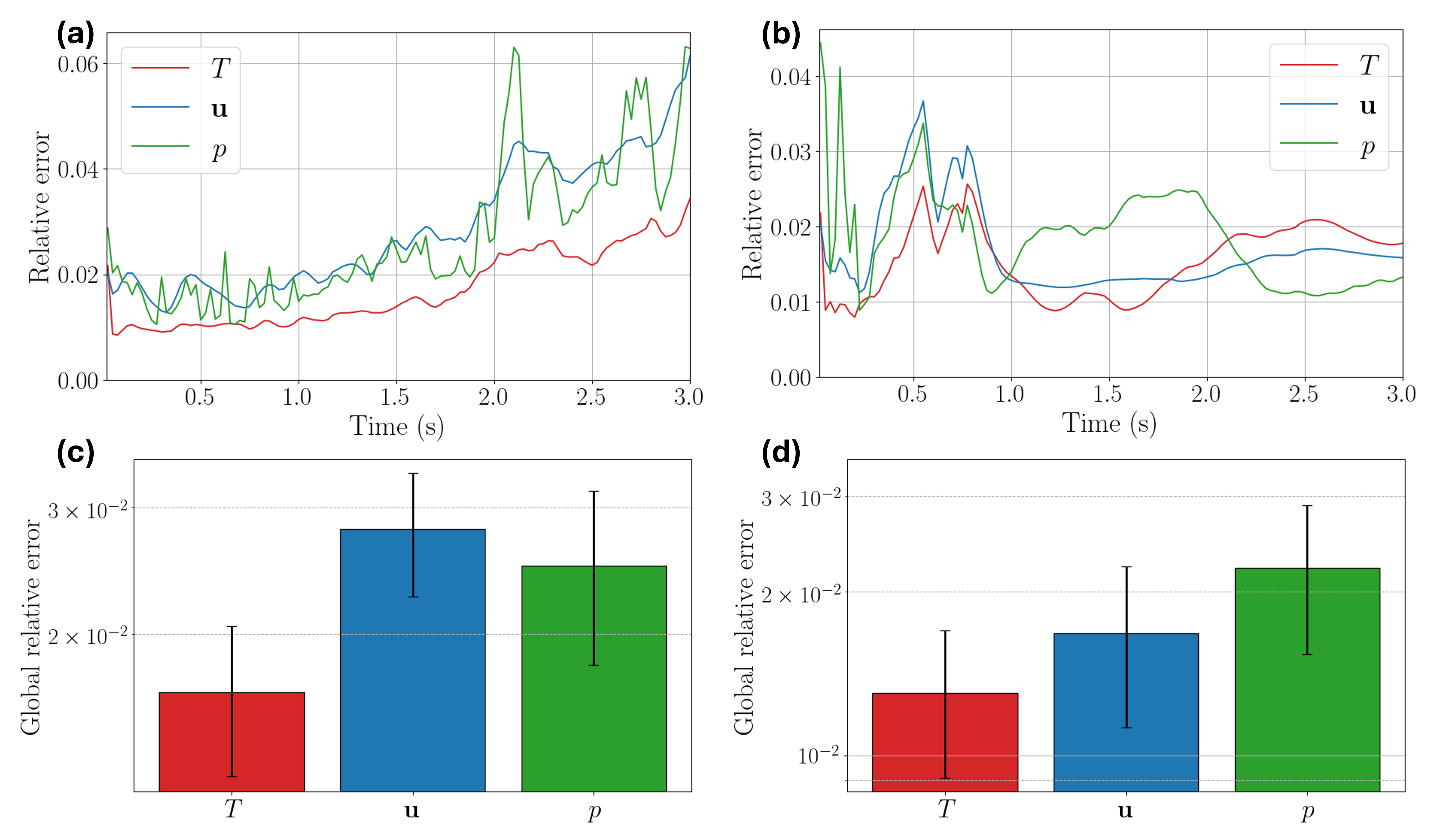}
    \caption{Temporal behavior of relative $L^2$-error of the SHRED reconstruction over time for temperature, velocity and pressure for the cases with $B_0 = 0.06 \ T$ (a) and $B_0 = 0.3 \ T$ (b). Global average over time of the relative error and related standard deviation (whiskers) for the cases with $B_0 = 0.06 \ T$ (c) and $B_0 = 0.3 \ T$ (d).}
    \label{fig: error}
\end{figure}

Moreover, the relative $L^2$-error related to the SHRED reconstruction over time has been calculated. For the given field $\psi$, the relative error has been computed as:
\begin{equation}
    \epsilon_\psi=\frac{\left\|\psi_{F O M}-\psi_{SHRED}\right\|}{\left\|\psi_{F O M}\right\|}
\end{equation}
where $\left\| \cdot \right\|$ represents the classical $L^2$-norm. Figure \ref{fig: error} shows the relative errors for both the considered test cases. 

In the scenario with low magnetic field intensity (Figure \ref{fig: error}-(a)), the reconstruction error exhibits a mild growth over time but remains consistently low throughout the entire time interval. Specifically, the error stays below approximately $6 \%$ for velocity and pressure and $3 \%$ for the temperature. The observed increase in error is attributable to the fact that the flow does not yet reach a steady or quasi-steady regime in the considered period; instead, the system continues to present dynamic evolution, as already shown by the temporal profile of the spatially averaged quantities (Figure \ref{fig: mean_field_lowB}), which keep varying and oscillating over time. Nevertheless, despite this gradual growth, the relative error remains very small for all fields. 

Furthermore, in the scenario with higher magnetic field intensity (Figure \ref{fig: error}-(b)), the error is even lower and follows a much more stable profile, eventually stabilizing at about  $2 \%$ for all physical fields. This behavior is fully consistent with the corresponding temporal evolution of the spatially averaged fields (Figure \ref{fig: mean_field_highB}), which tends toward a plateau within the considered time interval, since the flow approaches a more stationary regime under stronger magnetic influence. Moreover, the global mean relative error, obtained by averaging the relative error over time, and the related standard deviation have been computed for both the test cases (Figure \ref{fig: error}-(c) and \ref{fig: error}-(d)). They provide a global and cumulative assessment of the model accuracy over the entire time window, confirming the excellent overall performance of SHRED, since the mean error is below $3\%$ for all fields and the standard deviations remain small.

\section{Conclusions} \label{sec: Conclusions}

This work presents the first application of the SHallow REcurrent Decoder to magnetohydrodynamic physics involved in nuclear fusion reactors. The SHRED network was trained with scenarios involving a wide range of different magnetic field intensities. The results demonstrate that SHRED, once trained, starting only from the measure of the temperature in $3$ random points, is able to accurately reconstruct the flow dynamics (temperature, velocity and pressure) across the entire geometry for magnetic field intensities not seen during training, successfully reproducing flow regimes ranging from weakly magnetized and dynamically evolving configurations to strongly damped and fully laminarized flows. Moreover, the SHRED proved to be robust with respect to sensor placement. Indeed $30$ different randomly generated configurations of $3$ sensors were investigated, and the resulting reconstructions exhibit negligible variability, confirming that the model maintains high accuracy independently of sensor locations. All these results make SHRED particularly suitable for fusion applications. Firstly, accurate full-state reconstruction may be achievable by leveraging the intrinsic multiphysics of MHD flows, using only measurements of the temperature, which is the easiest and most practical quantity to access in fusion blankets. Secondly, sensors may be installed wherever they is most accessible, safe, or convenient, without requiring extensive optimization studies to determine ideal locations. This property is especially advantageous in fusion environments, where geometric constraints and extreme operating conditions may limit sensor placement. Furthermore, a single SHRED model, once trained over a broad range of magnetic fields, may be used to accurately reconstruct flow regimes that differ substantially from one another, capturing the distinct MHD effects that emerge under different magnetic configurations.

Overall, the presented methodology offers a computationally efficient, fully data-driven framework for real-time or multi-query MHD state reconstruction. Its ability to infer the full multiphysics state from sparse measurements highlights its potential for integration into monitoring, diagnostics, and control pipelines in fusion reactors. Future works will focus on extending the methodology to more complex and realistic geometries even in a three-dimensional framework. Moreover, the approach can be implemented in scenarios involving more complex magnetic configurations, such as time-varying profiles or magnetic fields with multiple spatial components.

\clearpage

\bibliography{bibliography.bib}

\end{document}